\documentclass[sigconf]{acmart}

\usepackage{booktabs} 
\usepackage{subfigure}
\usepackage{diagbox}
\usepackage{amsmath,bm}
\usepackage{setspace}  
\usepackage[ruled,vlined]{algorithm2e}
\usepackage{multirow}
\usepackage{makecell}
\usepackage{xcolor}
\usepackage{flushend}

\newcommand\given[1][]{\:#1\vert\:}
\DeclareMathOperator*{\argmax}{arg\,max}

\copyrightyear{2018} 
\acmYear{2018} 
\setcopyright{acmcopyright}
\acmConference[MM '18]{2018 ACM Multimedia Conference}{October 22--26, 2018}{Seoul, Republic of Korea}
\acmBooktitle{2018 ACM Multimedia Conference (MM '18), October 22--26, 2018, Seoul, Republic of Korea}
\acmPrice{15.00}
\acmDOI{10.1145/3240508.3240588}
\acmISBN{978-1-4503-5665-7/18/10}

\begin{document}
\title{GNAS: A Greedy Neural Architecture Search Method for Multi-Attribute Learning}

\author{Siyu Huang$^{1\text{\dag}}$, Xi Li$^{1*}$, Zhi-Qi Cheng$^{2\text{\dag}}$, Zhongfei Zhang$^1$, Alexander Hauptmann$^3$}
\affiliation{\institution{$^1$Zhejiang University, $^2$Southwest Jiaotong University, $^3$Carnegie Mellon University}}
\email{{siyuhuang,xilizju,zhongfei}@zju.edu.cn, zhiqicheng@gmail.com, alex@cs.cmu.edu}

\renewcommand{\shortauthors}{Siyu Huang et al.}

\begin{abstract}
A key problem in deep multi-attribute learning is to effectively discover the inter-attribute correlation structures. Typically, the conventional deep multi-attribute learning approaches follow the pipeline of manually designing the network architectures based on task-specific expertise prior knowledge and careful network tunings, leading to the inflexibility for various complicated scenarios in practice. Motivated by addressing this problem, we propose an efficient greedy neural architecture search approach (GNAS) to automatically discover the optimal tree-like deep architecture for multi-attribute learning. In a greedy manner, GNAS divides the optimization of global architecture into the optimizations of individual connections step by step. By iteratively updating the local architectures, the global tree-like architecture gets converged where the bottom layers are shared across relevant attributes and the branches in top layers more encode attribute-specific features. Experiments on three benchmark multi-attribute datasets show the effectiveness and compactness of neural architectures derived by GNAS, and also demonstrate the efficiency of GNAS in searching neural architectures.
\end{abstract}


\maketitle

\section{INTRODUCTION}
As an important variant of multi-task learning \cite{he2017adaptively}
and transfer learning \cite{yosinski2014transferable}, multi-attribute learning aims to discover the underlying correlation structures among attributes, which can improve the generalization performance of attribute prediction models by transferring and sharing information across multiple related attributes.
With the representation power of deep learning~\cite{rudd2016moon,hand2017attributes}, the problem of discovering such correlation structures is typically cast as designing tree-structured neural networks, whose architectures capture the attribute ontology properties in the forms of shared parent trunk networks followed by different child branch networks. Namely, the more semantically correlated attributes will share more parent trunk network layers followed by individual attribute-specific branch network layers. In this way, building an effective neural architecture is a key issue to solve in multi-attribute learning. 

\begin{figure}[t]
\includegraphics[width=0.91\linewidth]{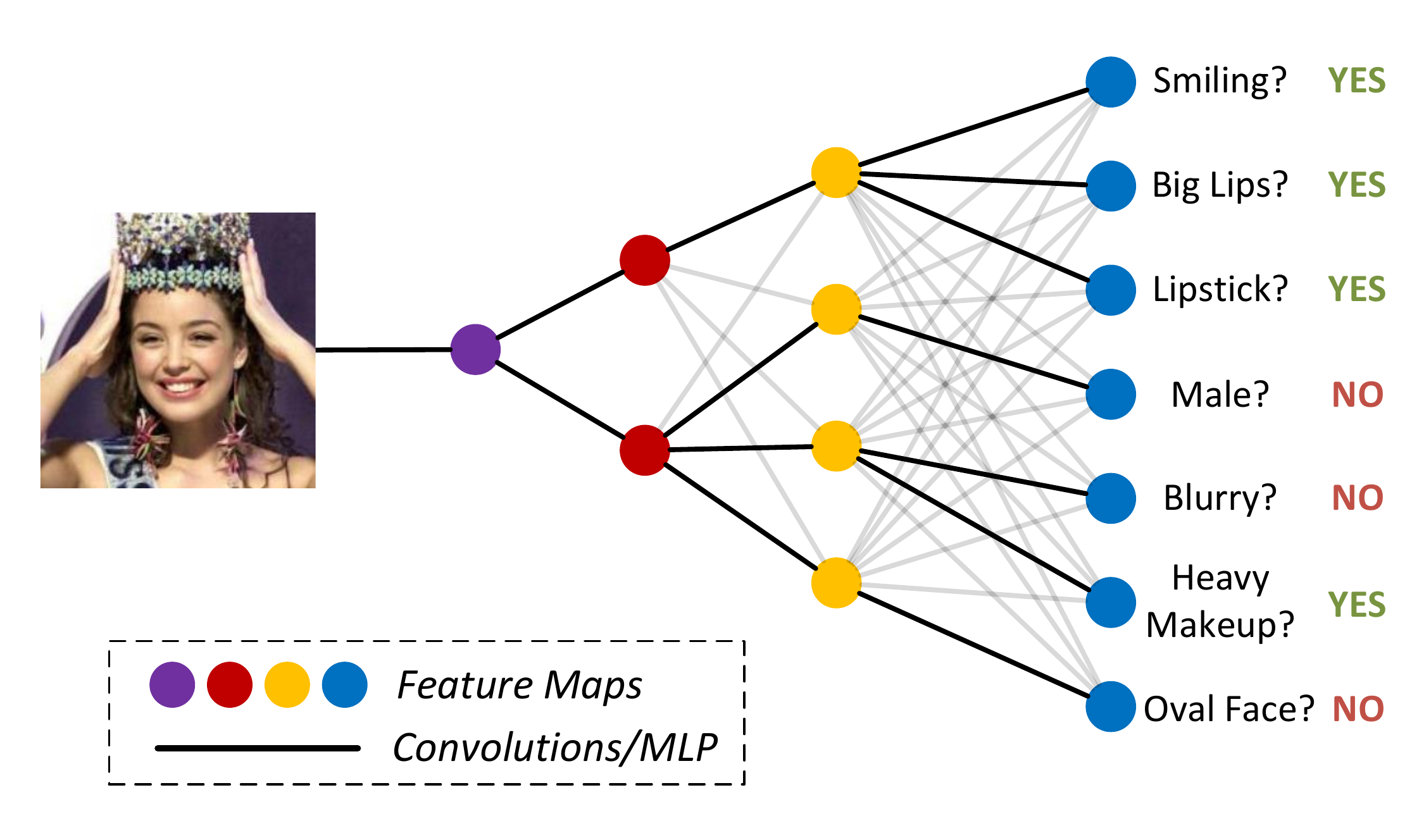}
\caption{Brief illustration of neural architecture search in multi-attribute learning. Our goal is to automatically discover the optimal tree-like neural network architecture from the combinatorially large space in order to jointly predict the attributes.
}
\label{fig1}
\end{figure}

Motivated by the above observations, a number of deep multi-attribute networks are built in a hand-designed way, which relies heavily on the expertise
knowledge in specific tasks. In practice, such a way is often heuristic,
inflexible, and incapable of well adapting to complicated real-world scenarios. 
In order to address this problem, we resort to automatically 
building the multi-attribute network architecture within an end-to-end learning
framework. As illustrated in Fig. \ref{fig1}, our goal is to discover the optimal tree-like architecture, where its root node is the input image and its leaf nodes are the probabilities of attributes. The low-level representations are more commonly shared and high-level representations are more task-specific, 
nicely fitting the nature of multi-attribute learning. However, it is a very challenging task to search architectures within such a combinatorially large space of possible connections. First, the number of candidate architectures is an exponential complexity of attribute numbers. For the example of Fig. \ref{fig1}, the number of candidate architectures of the last layer (between 4 yellow nodes and 7 blue nodes) is $4^7$=16,384. Second, it is computationally expensive to evaluate candidate architectures, as the evaluation has to be conducted after training a neural network to convergence.

In this paper, we propose a highly efficient greedy neural architecture 
search method (GNAS) to optimize the neural architecture for multi-attribute prediction. 
Inspired by the effective layer-wise pretraining strategy  \cite{hinton2006reducing,hinton2006fast,bengio2007greedy} 
proposed in earlier literature, we formulate the optimization of a global architecture as a series of sub-tasks of optimizing the independent layer architectures in a greedy manner. The optimization of a layer architecture 
is further divided into the optimizations of connections w.r.t individual attribute performance based on the property of tree structure. The optimal global architecture is derived by a combination of the optimal local  architectures after iteratively updating the local architectures and the neural network weights.

Our proposed GNAS approach is efficient and effective in the following aspects:
\begin{itemize}
\item
With the help of greedy strategies, GNAS reduces the number of candidate evaluated architectures from exponential complexity to linear complexity of the attribute number.
\item
GNAS could significantly accelerate the back propagation training of individual candidate architectures by incorporating the weight sharing mechanism \cite{real2017large,pham2018efficient} across different candidate architectures.
\item
GNAS could be used for searching arbitrary tree-structured neural network architecture. The large search space of GNAS ensures the performance of its discovered architecture.
\item
GNAS is a non-parametric approach that it refrains from the loop of adopting extra parameters and hyper-parameters for meta-learning (such as Bayesian optimization (BO) \cite{snoek2012practical} and reinforcement learning (RL) \cite{zoph2016neural,zoph2017learning}).
\end{itemize}

GNAS is not only theoretically reasonable, but also showing favorable performance in empirical studies. On three benchmark multi-attribute datasets, GNAS discovers network architectures on 1 GPU in no more than 2 days to beat the state-of-the-art multi-attribute learning methods with fewer parameters and faster testing speed.

The main contributions of this work are summarized as follows:
\begin{itemize}
\item
We propose an innovative greedy neural architecture search method (GNAS) 
for automatically learning the tree-structured multi-attribute deep network architecture. In principle, GNAS is efficient due to its greedy strategies, effective due to its large search space, and generalized due to its non-parametric manner.
\item
Experimental results on benchmark multi-attribute learning datasets demonstrate the effectiveness and compactness of deep multi-attribute model derived by GNAS. In addition, detailed empirical studies are conducted to show the efficacy of GNAS itself.
\end{itemize}

\section{RELATED WORK}
{\noindent \bf Multi-attribute learning.}
Similar to multi-task learning \cite{li2017multi}, multi-attribute learning addresses the attribute prediction problems by feature sharing and joint optimization across related attributes. In the context of deep attribute learning, prior works \cite{rudd2016moon,hand2017attributes,cao2018partially,liu2018deep} investigate designing end-to-end tree-like network architecture which shares feature representations in bottom layers and encode task-specific information in top layers. The tree-like architecture is able to improve the compactness and generalization ability of deep models.

However, the hand-designed network architecture raises a high demand of knowledges in specific tasks and experience in building neural networks. Motivated by this, researchers investigate the automatic design of deep architectures more recently. Cross-stitching network \cite{misra2016cross} is proposed to learn an optimal linear combination of shared representations, and \citeauthor{he2017adaptively} \cite{he2017adaptively} adaptively learn the weights of individual tasks. The work most close to our approach is \cite{lu2017fully} which first initializes a thin network from a pre-trained model by SOMP \cite{tropp2006algorithms} and then widening the network through a branching procedure. However, these approaches generally explore a relatively limited search space.

In this work, our proposed greedy neural architecture search method (GNAS) addresses the automatic design of deep multi-attribute architecture in an entirely different way. From the perspective of neural architecture optimization, GNAS divides the global architecture optimization problem into a series of local architecture optimization problems based on reasonable intra-layer and inter-layer greedy strategies. The greedy manner ensures the efficiency of architecture search procedure.

\begin{figure*}[t]
\includegraphics[width=1\linewidth]{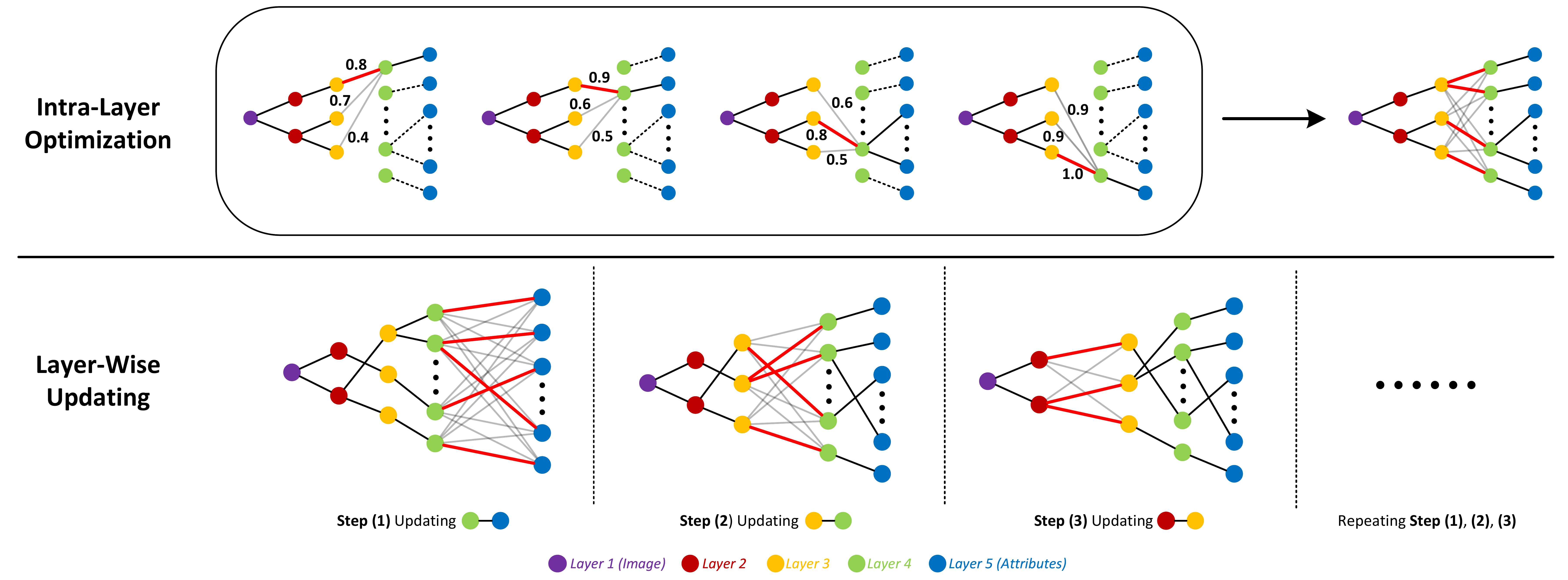}
\vspace{-0.5em}
\caption{Illustration of our greedy neural architecture search (GNAS). We transform the difficult global architecture optimization problem into a series of local architecture optimization problems. The upper part illustrates the optimization of intra-layer connections, where we respectively evaluate all the connections and select the connections which have the best validation performances on their descendant attributes. The lower part illustrates the layer-wise updating procedure, where we iteratively update the architecture of one layer conditioned on the fixed architectures of the other layers.
}
\label{gnas framework}
\end{figure*}

{\noindent \bf Neural architecture optimization.}
Deep neural network has achieved a great success on many tasks. While, the design of neural network architecture still relies on the expertise in neural network and prior knowledge of specific tasks. Recently, there is a growing amount of research focusing on the automatic design of neural network architecture, aiming at discovering the optimal neural architecture with less human involvement. A variety of approaches including random search \cite{bergstra2012random}, Bayesian optimization \cite{snoek2012practical,mendoza2016towards,kandasamy2018neural}, evolutionary algorithm \cite{real2017large}, and reinforcement learning \cite{zoph2016neural,pham2018efficient} are proposed for neural architecture optimization. The recently proposed neural architecture search (NAS) \cite{zoph2016neural,zoph2017learning} employs an RNN controller to sample candidate architectures and updating the controller under the guidance of performances of sampled architectures. Although models derived by NAS have shown impressive results on image classification and language modeling, the prohibitive expense of NAS limits its further development. As the learning of neural network is a black-box optimization, we have to evaluate an candidate neural architecture after it is trained to convergence. Typically, \citeauthor{zoph2016neural} \cite{zoph2016neural} use 800 GPUs and 28 days to discover the convolutional architecture on Cifar-10 dataset by exploring 12,800 individual architectures. 

Several approaches explore to accelerate the searching procedure by reducing the expense of neural network training. \citeauthor{baker2018accelerating} \cite{baker2018accelerating} early stop the architecture evaluation process by predicting the performance of unobserved architectures based on a set of architecture features. \citeauthor{brock2018smash} \cite{brock2018smash} propose a hypernetwork to generate the neural network weights conditioned on its architecture instead of conducting back propagation training. \citeauthor{pham2018efficient} \cite{pham2018efficient} search for an optimal sub-graph within a large computational graph where the neural network weights are shared across sub-graphs. 

In this work, we propose GNAS to novelly develop neural architecture optimization to multi-task learning. Different from existing neural architecture optimizing approaches, we propose two greedy strategies which largely reduce the computation cost of architecture optimization procedure. The intra-layer greedy strategy of GNAS is proposed based on the property of tree structure. And the inter-layer greedy strategy of GNAS is inspired by the layer-wise pretraining strategy of restricted Boltzmann machine (RBM) \cite{hinton2006reducing,hinton2006fast,bengio2007greedy}. The greedy strategies lead to the efficiency of GNAS, also leading to effectiveness by ensuring a highly efficient searching in a very large search space.


\section{OUR APPROACH}
\subsection{Problem Formulation}
Our goal is to find the optimal tree-like neural network architecture $\hat{G}$ which has the maximum reward $R$ 
\begin{align}
\label{general nas}
\hat{G} =  & \operatorname*{arg\,max}_G R(G) \\
= & \operatorname*{arg\,max}_G \frac{1}{N} \sum^N_{n=1} r_{n} (G) \notag
\end{align}
$R$ is defined as the mean prediction accuracy of attributes on validation set, where $r_{n}$ is the prediction accuracy of the $n$-th attribute on validation set and $N$ is the number of attributes. $G$ is the multi-output network with an input of an image and $N$ outputs for predicting $N$ attributes. $G$ is tree-like that it has $M$ layers. In each layer $l$, there are $B_l$ blocks where each block consists of a fixed number of feature maps. $B_1=1$ as the first layer is the input image and $B_M=N$ as the last layer is $N$ outputs of attribute predictions. $G$ hierarchically groups the related attributes from its top layers to bottom layers. 

For convenience, we use a set of binary adjacency matrices $A$ to denote the network topology of neural network $G$. $A^{(l)}_{i,j} = 1$ denotes that there is a connection (fixed as convolutions or MLP as needed) between the $i$-th block of layer $l$ and the $j$-th block of layer $l$+1, otherwise, $A^{(l)}_{i,j}=0$. We rewrite Eq. \ref{general nas} as 
\begin{align}
\label{attribute nas}
& \hat{A}= \operatorname*{arg\,max}_A R(A),  
& ~ \text{s.t.}~~  \sum^{B_l}_{i=1}{A^{(l)}_{i,j}}=1, \ 1 \leq j \leq B_{l+1} 
\end{align}
$A$ is constrained to be a tree structure under the constraint of Eq. \ref{attribute nas}. 

Eq. \ref{attribute nas} is a combinatorial optimization problem which has $\prod\limits_{l} B_{l}^{B_{l+1}}$ possible solutions. Therefore, it is often infeasible to get its optimal solution due to the large solution space. For instance, for a neural network with 40 output predictions and a hidden layer of 10 blocks, the number of possible neural architectures is $10^{40}$, such at we could not evaluate all of the possible architectures to find an optimal one. 

In this work, we present a non-parametric approach, i.e., GNAS, to search for the multi-output tree-like neural architecture effectively and efficiently. Generally speaking, we divide the global optimization problem into the optimization problems of individual layer architectures, and further dividing them into the optimization problems of individual connections. The optimal global architecture is approximated by the combination of optimal local architectures. More details of our approach are discussed in the following sections.

\subsection{Intra-Layer Optimization}
\label{Searching within a layer}
Our GNAS starts from optimizing the neural network connection w.r.t. an individual attribute within a layer. Given the architecture of the other layers, the problem is formulated as
\begin{equation}
\label{basic}
\operatorname*{arg\,max}_{A^{(l)}} r_n\left(A^{(l)} \given[\big] A^{(L)}, L \neq l\right),
\qquad \text{s.t.} \sum_{i,j}{A^{(l)}_{i,j}=1}
\end{equation}
Eq. \ref{basic} is easy to solve as our neural architecture is a tree structure, such that we only have to evaluate the connections between $B_l$ blocks of layer $l$ and the ancestor block of attribute $n$ in layer $l$+1.

To optimize the connections of an entire layer, we propose a greedy assumption:

{\noindent \bf Assumption 1}
{\it The optimal intra-layer architecture is composed by the optimal connections w.r.t. individual attributes.}

This assumption is definitely reasonable because our network structure is a tree. The connections from a block to its descendant attributes are unique thus the connections w.r.t. individual attributes in layer $l$ are nearly independently when connections of the other layers are fixed. Based on Assumption 1, we reformulate the optimization of a layer as optimizing a set of Eq. \ref{basic} independently,
\begin{align}
\label{in-layer}
& \argmax_{A^{(l)}} R\left(A^{(l)} \given[\big] A^{(L)}, L \neq l\right),
\qquad\qquad \text{s.t.} \sum_{i,j}{A^{(l)}_{i,j}}=N \\
=\ &  \argmax_{A^{(l)}} \frac{1}{N} \sum_{n=1}^{N} r_n\left(A^{(l)} \given[\big] A^{(L)}, L \neq l\right),
\quad \text{s.t.} \sum_{i,j}{A^{(l)}_{i,j}}=N \notag \\
\simeq \ &  \bigg\{ \argmax_{A^{(l)}} r_n\left(A^{(l)} \given[\big] A^{(L)}, L \neq l  \right), \quad \text{s.t.} \sum_{i,j}{A^{(l)}_{i,j}=1} \bigg\} \ \text{for}\ n=1,...,N \notag
\end{align}

Note that there may be more than one connections built from layer $l$ to a certain block of layer $l$+1 if $B_{l+1}$<$N$, leading to the destruction of the tree structure. To avoid this, we give each block an index $I^{(l)}_i \subseteq \{1,2,...,N\}$ denoting which attributes are the descendants of the $i$-th block of layer $l$. The network is tree-structured that the reward of a connection $A^{(l)}_{i,j}$ is exactly the average accuracy of its descendant attributes,
\begin{equation}
\label{block}
R\left(A^{(l)}_{i,j} \given[\big] A^{(L)}, L \neq l\right)= \frac{1}{\big\vert I^{(l+1)}_j \big\vert} \sum_{n \in I^{(l+1)}_j  } r_n \left( A^{(l)}_{i,j} \given[\big]  A^{(L)}, L  \neq l \right) 
\end{equation}

We optimize w.r.t. blocks instead of attributes, formulated as
\begin{equation}
\begin{split}
\label{in-layer1}
\text{Eq.} \ref{in-layer} \simeq \bigg\{ \argmax_{A^{(l)}} \frac{1}{\big\vert I^{(l+1)}_j \big\vert} &\sum_{n \in I^{(l+1)}_j  } r_n \left( A^{(l)} \given[\big]  A^{(L)}, L  \neq l \right) ,  
\ \\[-0.1cm] & \text{s.t.} \sum_{i,j}{A^{(l)}_{i,j}=1} \bigg\}  
 \ \text{for}\ j=1,...,B_{l+1}  
\end{split}
\end{equation}
Eq. \ref{in-layer1} is also easy to solve as we only have to evaluate $B_l$ architectures for optimizing a block. Until now, the architectures evaluated within a layer is reduced from $B_{l}^{B_{l+1}}$ to $B_{l} \cdot B_{l+1}$.

The upper part of Fig. \ref{gnas framework} illustrates a simple example of our searching process within a layer. In the example, we aim at optimizing the third layer of the neural architecture, i.e., the connections between yellow blocks and green blocks. The four sub-figures in the box respectively illustrate the optimizations w.r.t four green blocks. The connections with red lines are selected because they have higher rewards than the other candidate connections. Note that in the third sub-figure, the green block is the ancestor of two attributes, such that its reward is computed by averaging the validation accuracies of those two attributes. As shown in the upper right of Fig. \ref{gnas framework}, the optimal architecture of this layer is composed by the selected connections.

\subsection{Accelerating Intra-Layer Search}
\label{accelerating}

Although the architectures searched within a layer is reduced from $B_{l}^{B_{l+1}}$ to $B_{l} \cdot B_{l+1}$ by Eq. \ref{in-layer1}, the computing cost is still large. We propose to further decrease the number of evaluated architectures from $B_{l} \cdot B_{l+1}$ to $B_l$. In fact, Eq. \ref{block} indicates that we could get the reward of connection $A^{(l)}_{i,j}$ according to the accuracies of its descendant attributes. Therefore, we could evaluate the rewards of connection between a block in layer $l$ and all the blocks in layer $l$+1 simultaneously, as there is a unique path between a layer and a certain attribute in this case. 

\begin{figure}[t]
\includegraphics[width=1\linewidth]{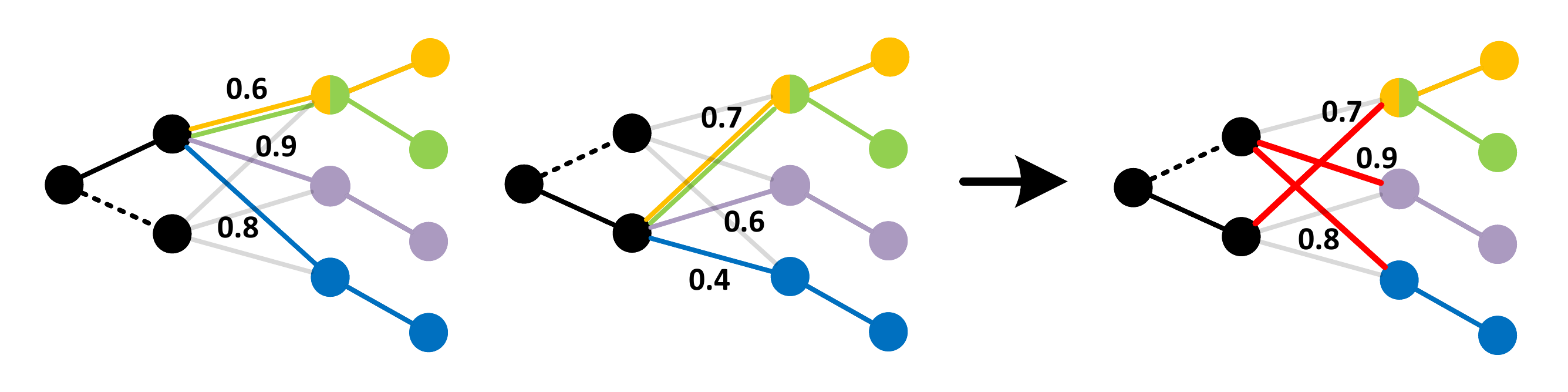}
\caption{Accelerating the intra-layer search by evaluating the connections between a black block and the colored blocks of the next layer at the same time. 
}
\label{parallel}
\end{figure}

As illustrated in Fig. \ref{parallel}, we aim at optimizing the connections between black blocks and colored blocks. We do not have to evaluate the possible connections separately, that is, we could evaluate the connections between a black block and all the colored blocks simultaneously. The reward of each connection comes from the validation accuracies of its descendant attributes. The connections with larger rewards are selected, as shown in the right of Fig. \ref{parallel}.

\subsection{Layer-Wise Updating}
To optimize the connections of the entire network, we have a greedy assumption: 

{\noindent \bf Assumption 2}
{\it The optimal global architecture is composed by the optimal layer architectures.}

This assumption is proposed based on the effective layer-wise pretraining strategy \cite{hinton2006reducing,hinton2006fast,bengio2007greedy} for initializing a restricted Boltzmann machine (RBM), where the weights of individual neural layers are separately pre-trained to ensure stable initialization. Similar to the neuron weights, the architectures of neural layers could also be viewed as the parameters of mapping functions. Thus, we propose Assumption 2 to layer-wise update the network architecture. The Eq. \ref{attribute nas} and Eq. \ref{in-layer} are connected as $\text{Eq.} \ref{attribute nas} \simeq \big\{ \ \text{Eq.} \ref{in-layer} \ \big\} \ \text{for} \ j=1,2,...,M-1$.
$M$ is the number of layers. As discussed in Section \ref{accelerating}, the number of evaluated architectures for updating a layer is $B_l$. Therefore, the number of evaluated architectures for optimizing the entire network is finally $\sum\limits_{l=1}^{M-1}B_l$. For instance a network with 4 layers of (1, 4, 16, 40) blocks in each layer, there are $\prod\limits_{l} B_{l}^{B_{l+1}}=6.28\times 10 ^{57}$ possible tree-structured architectures. By using our GNAS method, the number is decreased to $\sum\limits_{l=1}^{M-1}B_l=20$. 

The lower part of Fig. \ref{gnas framework} illustrates the layer-wise updating procedure. At every step, we update the connections of one layer while fixing the connections of the other layers based on Eq. \ref{in-layer1}. As the given condition $A^{(L)}, L  \neq l $ in Eq. \ref{in-layer1} will change after every update of the other layers, we repeat the layer-wise updating until convergence.

\begin{algorithm}[t]
\caption{Greedy neural architecture search (GNAS)}\label{alg}
\DontPrintSemicolon
\SetKw{Initialize}{Initialize}
\SetKwInOut{Input}{Input}
\SetKwInOut{Output}{Output}
\textbf{Input:} Training set $D_\text{train}$, validation set $D_\text{valid}$, layer number $M$, block number $B$\\
\textbf{Output:} Neural network architecture $A$ \\
\textbf{1. Initialization} \\
~~~~- Randomly initialize architecture $A$ subject to Eq. \ref{attribute nas}; \\
~~~~- Randomly initialize neural network weights $W$; \\
\textbf{2. Updating} \\
~~~~- \While {not converged}{
	~~~~-\For {$l$=$M$-1 downto 1}{
		~~~~- \For {$b$=1 to $B_l$}{
			~~~~- $ A^{(l)}_{i,j}\gets \left\{
					\begin{aligned}
					 1, & \ i=b \\
					 0, & \ i \neq b
					\end{aligned}
					\right. \ $; \\ 
			~~~~- Train $W[A]$ on batches of $D_\text{train}$; \\
			~~~~- $r(A)\gets$ Evaluate $W[A]$ on batches of $D_\text{valid}$; \\
		}
		~~~~- Update layer architecture $A^{(l)}$ based on $r$ by Eq. \ref{in-layer1};
	} 
}
\end{algorithm}  

{\noindent \bf Weight sharing.}
To evaluate the performance of a neural architecture, we have to take a long time to train it to convergence first. Thanks to the weight inheritance mechanism \cite{real2017large,pham2018efficient} proposed for neural architecture search, we share the weights of the same network connections across different architectures during the entire GNAS process. Specifically, we maintain the weights of network connections $A$ as $W[A]$. In training phase, the weight of connection $A^{(l)}_{i,j}$ is inherited from $W\left[A^{(l)}_{i,j}\right]$, and $W\left[A^{(l)}_{i,j}\right]$ is updated after training $A^{(l)}_{i,j}$. When evaluating $A^{(l)}_{i,j}$, the weight of $A^{(l)}_{i,j}$ is inherited from $W\left[A^{(l)}_{i,j}\right]$. We alternately train the network on several mini-batches of training set to update weights $W$, and evaluate the network on validation set to update architecture $A$, such that both the weights $W$ and the architecture $A$ get to convergence in this process. The complete GNAS algorithm is illustrated in Alg. \ref{alg}.

\section{Experiments}
\subsection{Implementation Details}
{\noindent \bf Datasets.}
In the experiments, we evaluate our approach on two facial attribute datasets: CelebA \cite{liu2015faceattributes} and LFWA \cite{LFWTech}, and one person attribute dataset: Market-1501 Attribute \cite{lin2017improving}. 
\begin{itemize}
\itemsep0em 
\item
\textbf{CelebA} dataset \cite{liu2015faceattributes} consists of 200k images respectively with 160k, 20k, and 20k images for training, validation, and testing sets. CelebA dataset and LFWA dataset provide the same 40 binary attributes. We randomly crop the images of CelebA to size (192, 168) for training.
\item
\textbf{LFWA} dataset \cite{LFWTech} consists of 13,143 images respectively with 6,263 and 6,880 for training and testing sets. As there is no official split of training and validation, we use the first 5,000 images in training set for training and the rest 1,263 images for validation. We randomly crop the images of LFWA to size (224, 224) for training.
\item
\textbf{Market-1501 Attribute} dataset \cite{lin2017improving} annotates 23 person attributes on the original Market-1501 dataset \cite{zheng2015scalable}. It has 32,688 images of 1501 identities, including 16,522 images of 751 identities for training and 17661 images of 750 identities for testing respectively. As there is no official split of training and validation, we use the first 13,000 images in training set for training and the rest 3,522 images for validation. 
\end{itemize}

Standard image preprocessing including normalization and random horizontal flip is applied to all the three datasets.

%

{\noindent \bf Network architecture.}
In this work, we propose GNAS to search for the optimal tree-structured neural network architecture which is a sub-graph of a pre-defined graph. In the experiments, we evaluate GNAS with several different configurations. As described in Table. \ref{network architecture}, we use two versions of pre-defined graphs, including GNAS-Shallow and GNAS-Deep. GNAS-Shallow searches for connections within layers of \textit{Conv-4}, \textit{Conv-5}, and \textit{FC-1}. GNAS-Deep searches for connections within layers of \textit{Conv-2, 3, 4, 5}, and \textit{FC-1}. $N$ is the attribute number corresponding to different datasets. We do not update the last two FC layers, i.e., they are fixed for attribute regression. For each graph version, we also employ two versions of channel numbers including Thin and Wide, where Thin version has fewer channels of feature maps and Wide version has more channels of feature maps in every layer. After each convolutional layer, we adopt a Batch Normalization (BN) layer, an ReLU layer, and a max-pooling layer with a kernel size of 2$\times2$ and a stride of 2. BNs are removed in inference phase for faster computing. Binary cross entropy loss is adopted at the output ends of the network to measure binary attribute predictions.

\begin{table}[t]
\centering
\small
\caption{Network Architecture Configuration}
\begin{tabular}{ccccccccc}
\toprule
\multirow{3}{*}{Layer} & \multirow{3}{*}{Kernel} &  \multicolumn{3}{c}{\textbf{Shallow}} & \multicolumn{3}{c}{\textbf{Deep}} \\
\cmidrule(lr){3-5} \cmidrule(lr){6-8} && \multirow{2}{*}{Block} & \multicolumn{2}{c}{Channel}  & \multirow{2}{*}{Block} & \multicolumn{2}{c}{Channel} \\ 
 \cmidrule(lr){4-5} \cmidrule(lr){7-8} &&& Thin & Wide && Thin & Wide \\
\midrule
Conv-1  & 7$\times$7 & 1  & 16 & 64 &  1  & 16 & 64 \\ 
Conv-2  & 3$\times$3 & 1  & 32 & 128 & \textbf{2}  & 16 & 64\\ 
Conv-3  & 3$\times$3 & 1  & 64 & 256 & \textbf{4}  & 16 & 64\\ 
Conv-4  & 3$\times$3 & \textbf{4}  & 32 & 128 & \textbf{8}  & 16 & 64\\ 
Conv-5  & 3$\times$3 & \textbf{16} & 16 & 32 & \textbf{16}  & 16 & 32\\  
FC-1    & -         & \textit{\textbf{N}} & 64 & 128& \textit{\textbf{N}}  & 64 & 128\\ 
FC-2    & -         & $N$  & 64 & 128& $N$  & 64 & 128\\ 
FC-3    & -          & $N$  & 2 & 2& $N$  & 2 & 2\\ 
\bottomrule
\end{tabular}
\label{network architecture}
\end{table}

\begin{table*}[t]
\centering
\small
\caption{Comparison with State-of-the-Art Facial Attribute Learning Methods}
\begin{tabular}{cccccc}
\toprule
\multirow{2}{*}{\textbf{Method}} & \multicolumn{2}{c}{\textbf{Mean Error} (\%)} 
& \multirow{2}{*}{\makecell{\textbf{Params} \\ (million)}} & \multirow{2}{*}{\makecell{\textbf{Test Speed} \\ (ms)}} & \multirow{2}{*}{\textbf{Adaptive?}} \\ 
\cmidrule(lr){2-3} & \textbf{CelebA} & \textbf{LFWA} \\
\midrule
LNets+ANet \cite{liu2015faceattributes} & 13 & 16 & - & - & No\\
Separate Task \cite{rudd2016moon} & 9.78 & - & - & -  & No\\
MOON \cite{rudd2016moon} & 9.06 & - & 119.73 & 12.53  & No\\
Independent Group \cite{hand2017attributes} & 8.94 & 13.72 & - & - & No\\
MCNN \cite{hand2017attributes} & 8.74 & 13.73 & - & -& No\\
MCNN-AUX \cite{hand2017attributes} & 8.71 & 13.69 & - & -& No\\
VGG-16 Baseline \cite{lu2017fully} & 8.56 & - & 134.41 & 12.60 & No\\
Low-rank Baseline \cite{lu2017fully} & 9.12 & - & 4.52 & 6.07 & No\\
\midrule
SOMP-thin-32 \cite{lu2017fully}  & 10.04 & - & 0.22 & 1.94 & Yes\\
SOMP-branch-64 \cite{lu2017fully} & 8.74 & - & 4.99 & 5.77 & Yes\\
SOMP-joint-64 \cite{lu2017fully} & 8.98 & - & 10.53 & 6.18 & Yes\\
PaW-subnet \cite{ding2017deep} & 9.11 & - & 0.27 & - & Yes\\
PaW \cite{ding2017deep} & 8.77 & - & 11 & - & Yes\\
\midrule
GNAS-Shallow-Thin & 8.70 & 13.84 & 1.57 & 0.33 & Yes\\
GNAS-Shallow-Wide & \textbf{8.37} & \textbf{13.63} & 7.73 & 0.64 & Yes\\
GNAS-Deep-Thin & 9.10 & 14.12  & 1.47 & 0.87 & Yes\\
GNAS-Deep-Wide & 8.64 & 13.94  & 6.41 & 0.89 & Yes\\
\bottomrule
\end{tabular}
\label{compare sota}
\end{table*}

{\noindent \bf Learning configurations.}
The deep neural networks are implemented based on PyTorch \cite{paszke2017automatic} in our experiments. For the training of neural networks, we use SGD with the learning rate of 0.1, the batch size of 64, the weight decay of $10^{-4}$, and the Nesterov momentum \cite{nesterov1983method} of 0.9. We train a sub-graph for 2000 iterations on CelebA and 400 iterations on LFWA and Market-1501 every time. The learning rate is decayed by 0.96 after a round of layer-wise updating. We fine-tune the selected architecture on both of the training sets and the validation sets, then reporting its performance on the testing set. As the training on LFWA and Market-1501 may easily overfit to training set due to their small number of samples, we adopt a Dropout layer \cite{srivastava2014dropout} with a drop rate of 0.75 after each fully-connected layer for LFWA and Market-1501. 

{\noindent \bf Running costs.}
On CelebA, the neural network weights and network architecture get converged after 150 rounds of layer-wise updating, taking about 2 days on a GTX 1080Ti GPU. On LFWA and Market-1501, the weights and architecture get converged after 300 rounds of layer-wise updating, taking about 1 days on a GTX 1080Ti GPU. 

\subsection{Multi-Attribute Prediction}
{\noindent \bf Facial attribute prediction.}
Table \ref{compare sota} compares our method with the state-of-the-art facial attribute prediction methods. The first group of methods design the model architectures by hand-craft, and the second group of methods derive the model architectures from data, as denoted by the column of `Adaptive?' in Table \ref{compare sota}. The testing speeds of the other methods are cited from \cite{lu2017fully}. As \cite{lu2017fully} uses a Tesla K40 GPU (4.29 Tflops) and we use a GTX 1080Ti GPU (11.3 Tflops), we convert the testing speeds of their paper according to GPU flop number. In addition, we use a batch size of 32 in testing for a fair comparison with \cite{lu2017fully}. 

Table \ref{compare sota} shows that our GNAS models outperform the other state-of-the-art methods on both of CelebA and LFWA datasets, with faster testing speed, relatively fewer model parameters, and feasible searching costs (no more than 2 GPU-days). It demonstrates the effectiveness and efficiency of GNAS in multi-attribute learning. The fast testing speed of GNAS model is mainly due to its fewer convolution layers (5 layers) and tree-like feature sharing architecture. Comparing different models derived by GNAS, GNAS-Shallow models perform better than GNAS-Deep models with faster speed and almost equal number of parameters, indicating that it is better to share high-level convolutional feature maps for multi-attribute learning. GNAS-Wide models perform better than GNAS-Thin models with the reason of employing more model parameters.

\begin{table}[t]
\centering
\small
\caption{Comparison of Person Attribute Learning Methods}
\label{person attribute}
\begin{tabular}{cc}
\toprule
\textbf{Method} & \textbf{Market-1501} (\%)  \\ 
\midrule
Ped-Attribute-Net \cite{lin2017improving} & 13.81 \\
Separate Models	\cite{he2017adaptively}  & 13.32	\\
APR \cite{lin2017improving} & 11.84	\\
Equal-Weight \cite{zhang2014panda} & 13.16	\\
Adapt-Weight \cite{he2017adaptively} & 11.51	\\
\midrule
\midrule
Random-Thin & 11.94\\
Random-Wide & 11.42	\\
\midrule
GNAS-Thin & 11.37	\\
GNAS-Wide & \textbf{11.17}	\\
\bottomrule
\end{tabular}
\end{table}

{\noindent \bf Person attribute prediction.}
Table \ref{person attribute} compares GNAS with the state-of-the-art person attribute learning methods. We only test our GNAS-Shallow-Thin and GNAS-Shallow-Wide , as Market-1501 Attribute dataset \cite{lin2017improving} has fewer attributes (27 binary attributes). We also test the random architecture including Random-Thin and Random-Wide which have the same numbers of blocks and channels corresponding to GNAS-Thin and GNAS-Wide. Table \ref{person attribute} shows that GNAS-Wide still performs the best compared to other methods including the state-of-the-art methods and the random baselines. The Adapt-Weight \cite{he2017adaptively} is also the adaptive method which adaptively learns the weights of tasks from data. Our method performs a little better, possibly due to the flexibility of GNAS-based models. GNAS-Thin and GNAS-Wide respectively outperform their random baselines by 0.57\% and 0.25\%, denoting the effectiveness of architectures derived by GNAS.

\begin{figure*}[!htb]
\centering
\subfigure{
\begin{minipage}[b]{0.45\linewidth}
\includegraphics[width=1\linewidth]{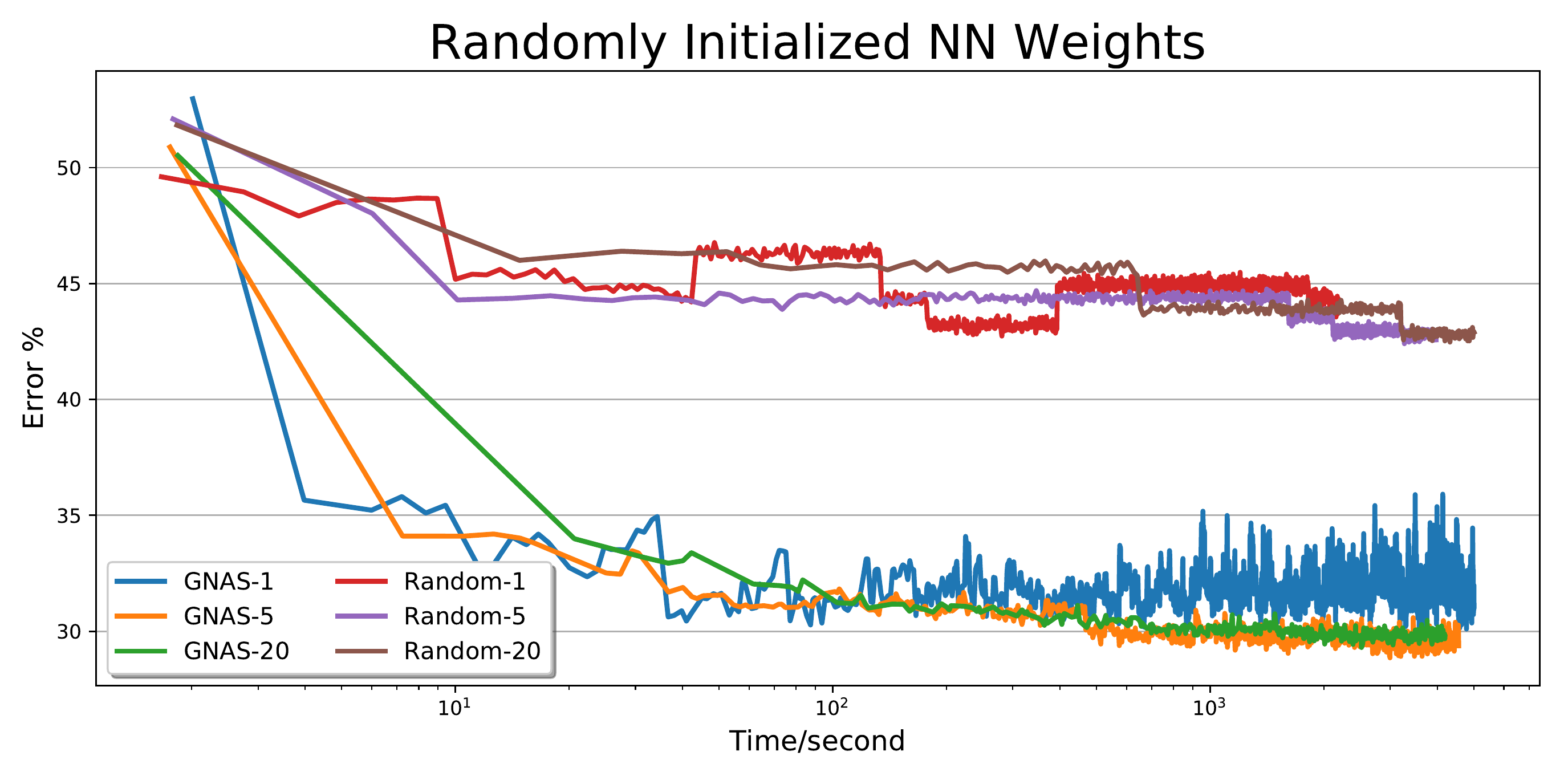}
\end{minipage}
}
\subfigure{
\begin{minipage}[b]{0.45\linewidth}
\includegraphics[width=1\linewidth]{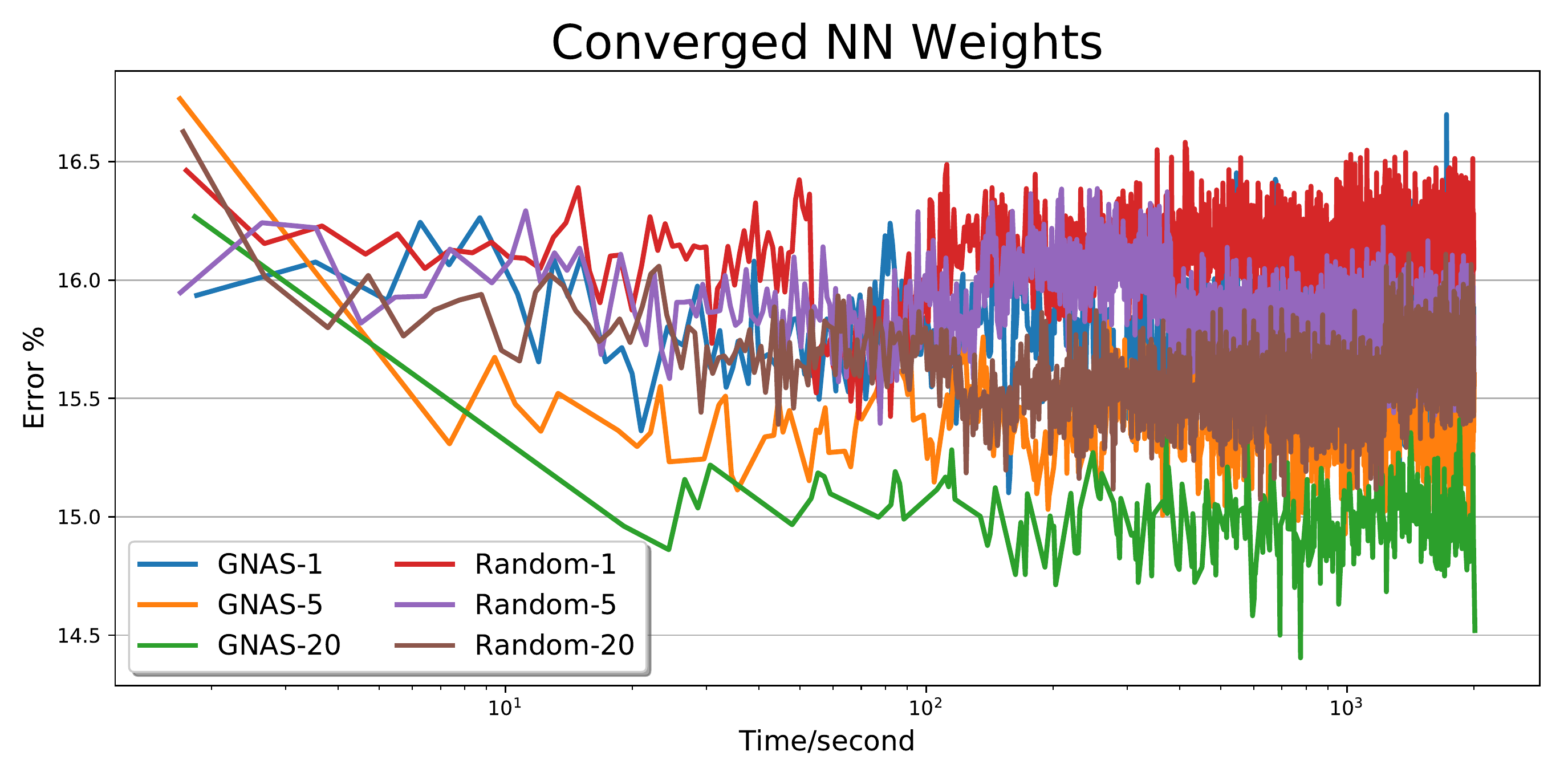}
\end{minipage}
}
\vspace{-0.5em}
\caption{Architecture search experiments on neural networks of random initialized weights and nearly converged weights. We compare random search and GNAS with different numbers of validation samples \{1, 5, 20\}. GNAS methods significantly outperform the random search methods with better performance and faster convergence speed. More validation samples contributes to better performance. Best viewed in color.
}
\label{compare valid}
\end{figure*}

\begin{figure}[t]
\centering
\subfigure{
\begin{minipage}[b]{0.54\linewidth}
\includegraphics[width=1\linewidth]{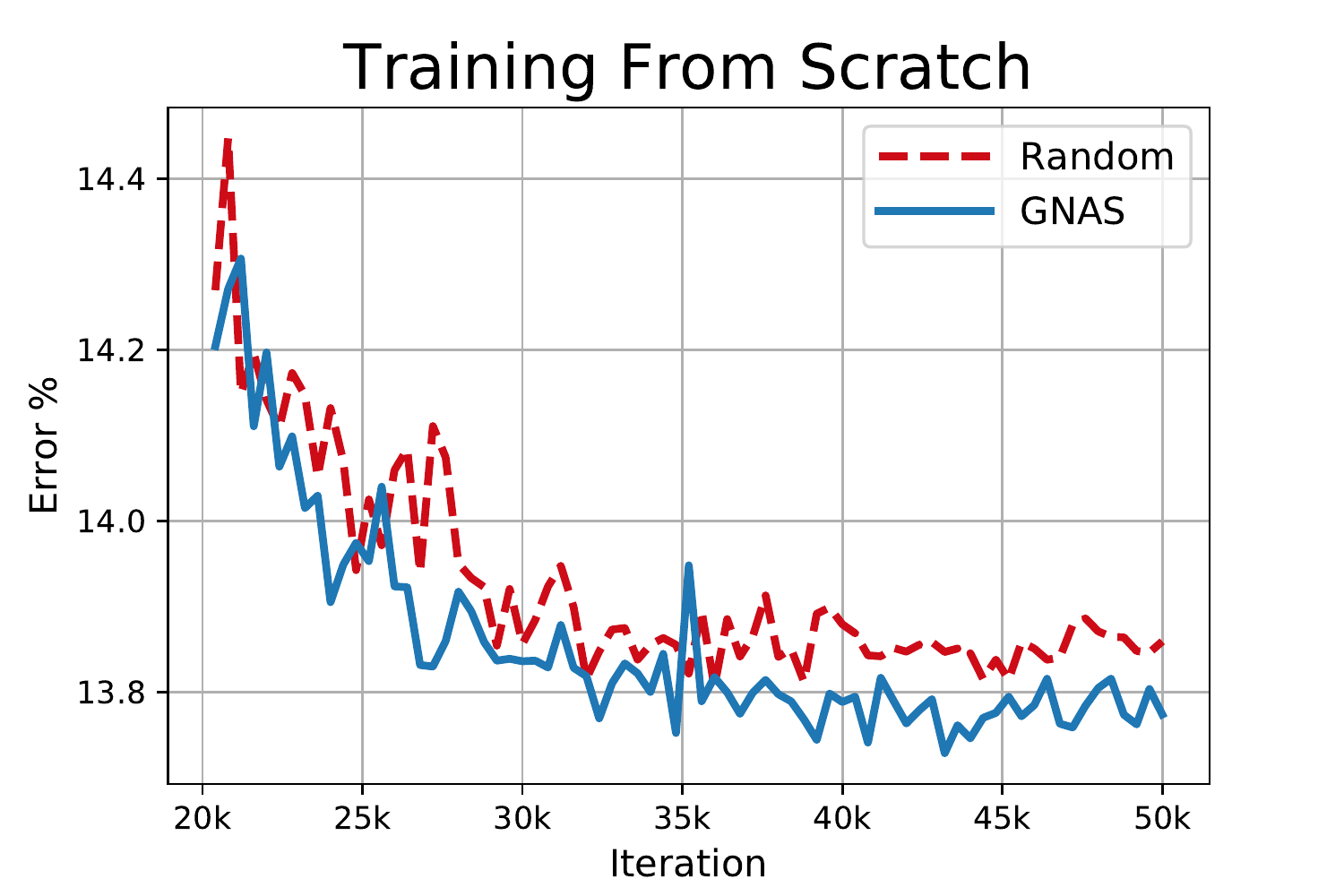}
\end{minipage}
}
\caption{Testing errors of models derived by random search and GNAS with training from scratch on LFWA. GNAS model converges faster and better.
}
\label{compare test}
\end{figure}

\subsection{Efficiency of GNAS}
It is known that the random search method is a strong baseline for black-box optimization \cite{bergstra2012random,zoph2017learning}. To demonstrate the effectiveness and efficiency of GNAS, we conduct more empirical studies on GNAS and random search. 

A good neural architecture search method should firstly be able to find the architecture performing good on validation set. Fig. \ref{compare valid} shows the performances of architectures discovered by random search and GNAS on the validation set of LFWA dataset, along with the logarithmic time scale. In random search, we randomly sample the neural architectures and output the one which has the best validation performance in history. The numbers in legends are the number of mini-batches used for evaluating. For instance, GNAS-1 denotes that we evaluate the reward of an architecture on 1 mini-batches at a time. 

In the left part of Fig. \ref{compare valid}, we randomly initialize the weights $W$ of neural network and make $W$ fixed during the searching process. GNAS outperforms random search by large margin in this case. Starting from the randomly initialized architecture which has about 50\% error rate on validation set, random search decreases the error rate to 43\% in one hour, while, GNAS could decrease the error rate to 30\% in fewer than 400 seconds. In addition, the number of validation samples has a significant impact on the performance of architecture search methods. Random-5 performs better than Random-20 at the beginning, while Random-20 shows a better performance after enough long time. The error rate of Random-1 even increases at some time because of the larger variance brought by its fewer validation samples. Similarly, GNAS-1 has larger mean and variance of error rate than those of GNAS-5 and GNAS-20. GNAS-5 and GNAS-20 show similar performance, indicating that 5 mini-batches of validation samples are sufficient for GNAS in this case.

In the right part of Fig. \ref{compare valid}, we inherit the neural network weights $W$ from a well-trained neural network and also fix $W$ during the searching process. Compared to the left part of Fig. \ref{compare valid}, the error rates of different methods are closer to each other in the right part of Fig. \ref{compare valid}. While, it is distinct that GNAS-20 performs the best and GNAS-5 performs the second-best. It demonstrates that GNAS could find better architecture than random search at different stages of the neural network training procedure. In addition, it reminds that GNAS should reduce its variance when searching architecture on a well-trained neural network by employing more validation samples. 

We also evaluate the performances of architectures derived by GNAS and random baseline on the testing set. As shown in Fig. \ref{compare test}, we train from scratch the architectures on LFWA dataset. The testing error rates of GNAS model and random baseline model are respectively shown as the solid line and the dashed line. The GNAS model performs better than the random baseline model with faster convergence speed and lower error rate. In summary, both the empirical results on validation set and testing set reveal the effectiveness and efficiency of our GNAS.

\begin{figure}[t]
\includegraphics[width=0.8\linewidth]{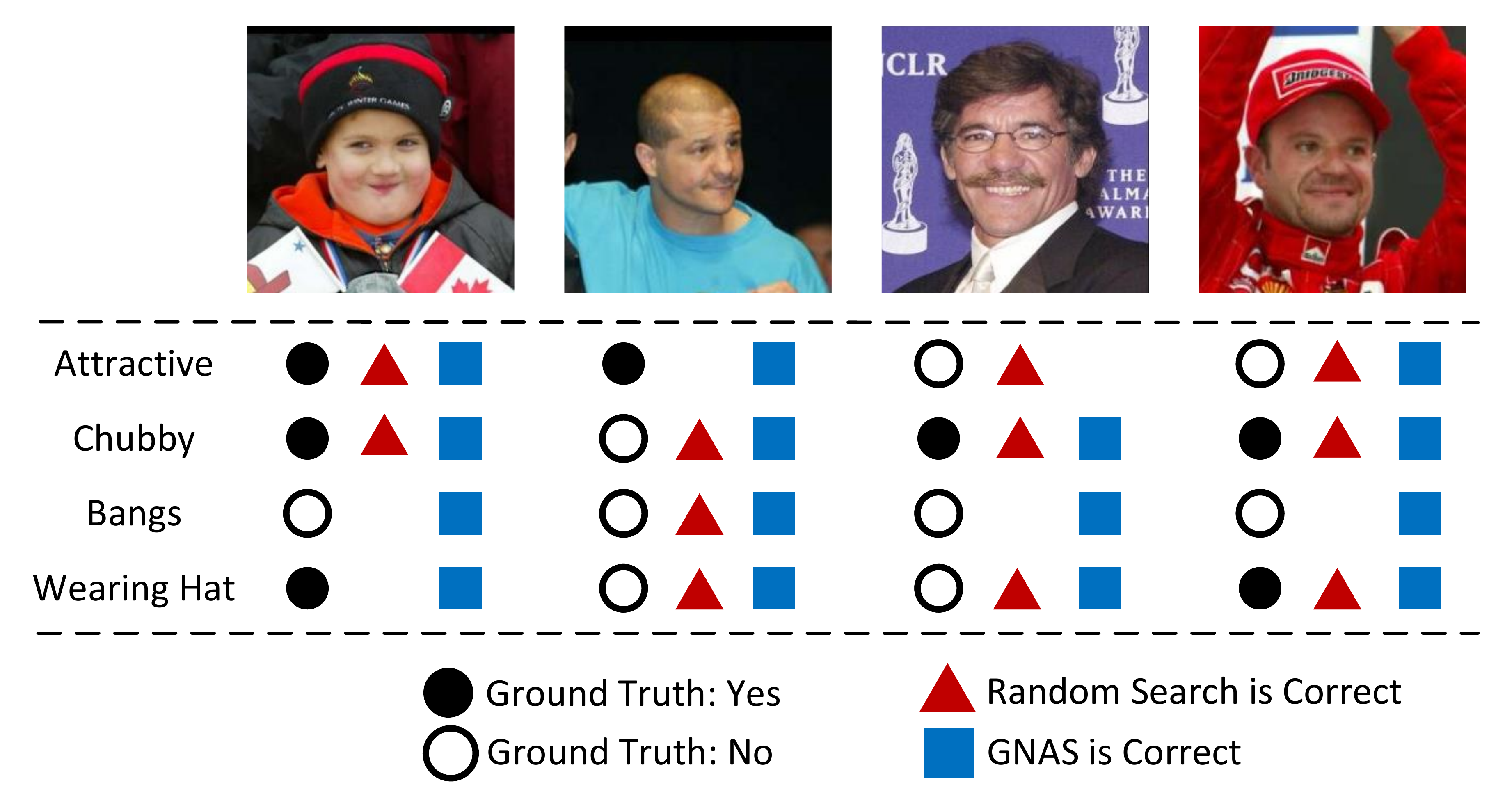}
\vspace{2pt}
\caption{Some qualitative results of LFWA dataset. GNAS model makes more correct predictions than random search model.
}
\label{qualitative}
\end{figure}

We further show some qualitative results in Fig. \ref{qualitative}. In Fig. \ref{qualitative}, the images and attributes come from the testing set of LFWA dataset. The ground truth annotations of `Yes' and `No' are respectively denoted by solid circles and hollow circles. The correct predictions of random baseline model and GNAS model are respectively denoted by triangles and squares. It is clear that GNAS performs better than random search in most of the cases. GNAS makes only an incorrect prediction on `Attractive' attribute of the third image, possibly due to the subjectivity of `Attractive'. These qualitative results also reveal the effectiveness of GNAS.

\subsection{Study on Attributes}

\begin{table}[t]
\centering
\scriptsize
\caption{Per-Attribute Performances on CelebA Dataset}
\label{per-attribute}
\begin{tabular}{lcccccc}
\toprule
\diagbox{Attribute}{Method} & LANet & Inde. & MCNN & M-AUX & PaW & GNAS \\ 
\midrule
5'o Clock Shadow & 9.00 & 6.06 & 5.59 & 5.49 & 5.36 & \textbf{5.24} \\
Arched Eyebrows & 21.00 & 16.84 & 16.45 & 16.58 & 16.99 & \textbf{15.75} \\
Attractive & 19.00 & 17.78 & 17.06 & \textbf{16.94} & 17.14 & \textbf{16.94} \\
Bags Under Eyes & 21.00 & 15.17 & 15.11 & 15.08 & 15.42 & \textbf{14.13} \\
Bald & 2.00 & 1.15 & 1.13 & 1.10 & 1.07 & \textbf{1.04} \\
Bangs & 5.00 & 4.01 & 3.96 & 3.95 & 4.07 & \textbf{3.80} \\
Big Lips & 32.00 & 29.20 & 28.80 & 28.53 & 28.54 & \textbf{28.21} \\
Big Nose & 22.00 & 15.53 & 15.50 & 15.47 & 16.37 & \textbf{14.90} \\
Black Hair & 12.00 & 10.59 & 10.13 & 10.22 & 10.16 & \textbf{9.76} \\
Blond Hair & 5.00 & 4.12 & 4.03 & 3.99 & 4.15 & \textbf{3.89} \\
Blurry & 16.00 & 3.93 & 3.92 & 3.83 & 3.89 & \textbf{3.58} \\
Brown Hair & 20.00 & 11.25 & 11.01 & 10.85 & 11.50 & \textbf{10.25} \\
Bushy Eyebrows & 10.00 & 7.13 & 7.20 & 7.16 & 7.38 & \textbf{7.01} \\
Chubby & 9.00 & 4.45 & 4.34 & 4.33 & 4.54 & \textbf{4.07} \\
Double Chin & 8.00 & 3.57 & 3.59 & 3.68 & 3.74 & \textbf{3.52} \\
Eyeglasses & 1.00 & 0.33 & 0.37 & 0.37 & 0.41 & \textbf{0.31} \\
Goatee & 5.00 & 2.87 & 2.70 & 2.76 & 2.62 & \textbf{2.41} \\
Gray Hair & 3.00 & 1.93 & 1.80 & 1.80 & 1.79 & \textbf{1.63} \\
Heavy Makeup & 10.00 & 9.05 & 8.63 & 8.45 & 8.47 & \textbf{8.18} \\
High Cheekbones & 12.00 & 12.66 & 12.45 & 12.42 & 12.56 & \textbf{11.95} \\
Male & 2.00 & 1.98 & 1.84 & 1.83 & 1.61 & \textbf{1.50} \\
Mouth Slightly Open & 8.00 & 6.01 & 6.26 & 6.26 & 5.95 & \textbf{5.84} \\
Mustache & 5.00 & 3.33 & 3.07 & 3.12 & 3.10 & \textbf{2.97} \\
Narrow Eyes & 19.00 & 12.78 & 12.84 & 12.77 & 12.44 & \textbf{12.34} \\
No Beard & 5.00 & 4.07 & 3.89 & 3.95 & 3.78 & \textbf{3.70} \\
Oval Face & 34.00 & 25.30 & 24.19 & \textbf{24.16} & 24.97 & 24.43 \\
Pale Skin & 9.00 & 2.93 & 2.99 & 2.95 & 2.92 & \textbf{2.76} \\
Pointy Nose & 28.00 & 22.53 & 22.53 & 22.53 & 22.65 & \textbf{21.76} \\
Receding Hairline & 11.00 & 6.59 & 6.19 & 6.19 & 6.56 & \textbf{6.06} \\
Rosy Cheeks & 10.00 & 4.98 & 4.87 & \textbf{4.84} & 4.93 & 4.99 \\
Sideburns & 4.00 & 2.23 & 2.18 & 2.15 & 2.36 & \textbf{2.04} \\
Smiling & 8.00 & 7.35 & 7.34 & 7.27 & 7.27 & \textbf{6.76} \\
Straight Hair & 27.00 & 17.38 & 16.61 & 16.42 & 16.48 & \textbf{15.23} \\
Wavy Hair & 20.00 & 16.76 & 16.08 & 16.09 & 15.93 & \textbf{15.48} \\
Wearing Earrings & 18.00 & 9.65 & 9.68 & 9.57 & 10.07 & \textbf{9.02} \\
Wearing Hat & 1.00 & 1.03 & 0.96 & 0.95 & 0.98 & \textbf{0.88} \\
Wearing Lipstick & 7.00 & 6.20 & 6.05 & 5.89 & 5.76 & \textbf{5.59} \\
Wearing Necklace & 29.00 & 13.59 & 13.18 & 13.37 & 12.30 & \textbf{12.39} \\
Wearing Necktie & 7.00 & 3.29 & 3.47 & 3.49 & \textbf{3.15} & 3.24 \\
Young & 13.00 & 12.02 & 11.70 & 11.52 & 11.41 & \textbf{11.11} \\ \midrule
Ave. & 12.67 & 8.94 & 8.74 & 8.71 & 8.77 & \textbf{8.37} \\
\bottomrule
\end{tabular}
\end{table}

\begin{figure}[t]
\centering
\subfigure{
\begin{minipage}[b]{0.8\linewidth}
\includegraphics[width=1\linewidth]{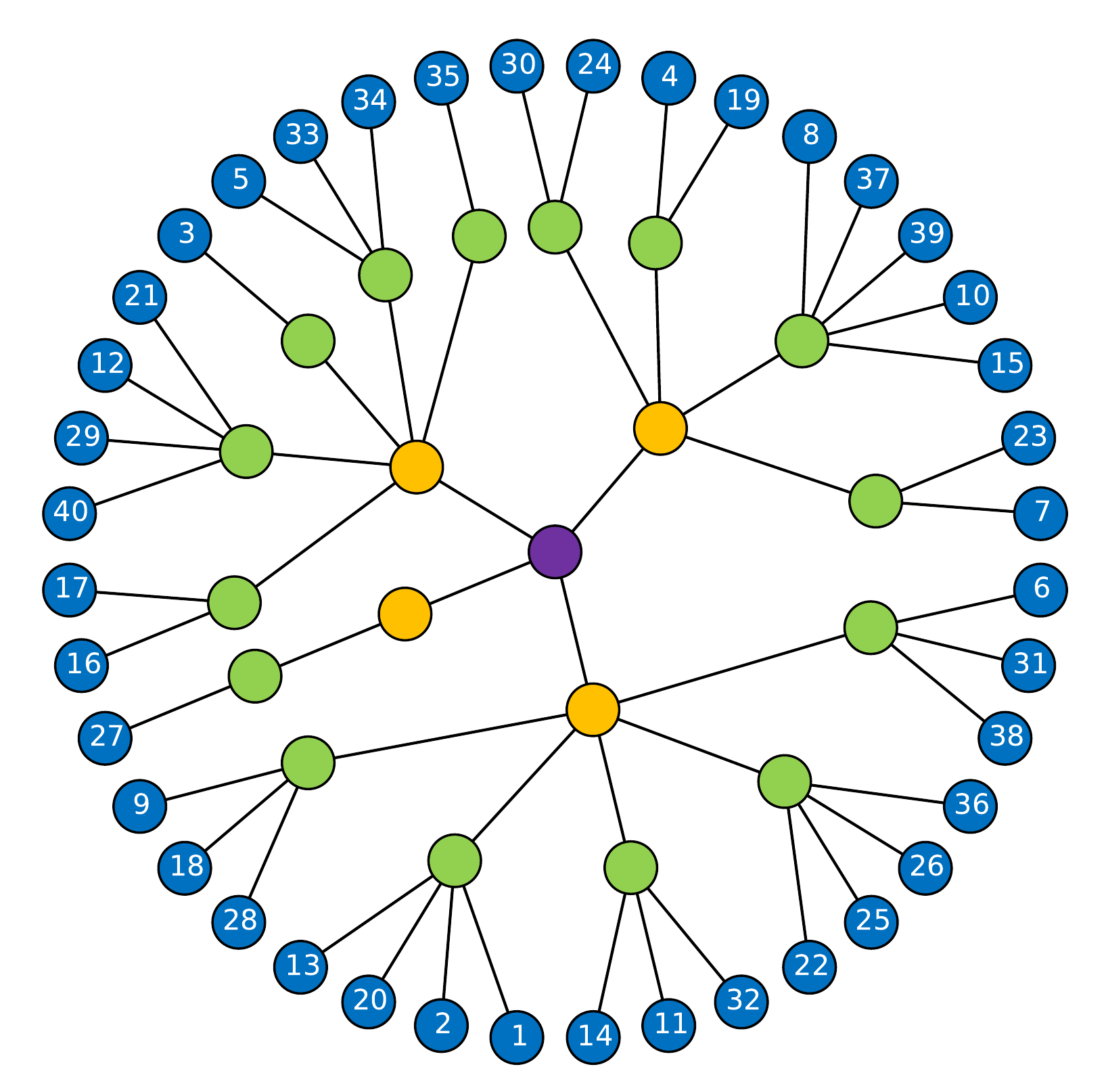}
\end{minipage}
}
\caption{Network architecture discovered by GNAS-Shallow-Thin on LFWA, where related attributes are hierarchically grouped together.
The 40 attributes are:
1 {\color{red}{\it 5'o Clock Shadow}} 2 {\color{red}{\it Arched Eyebrows}} 3 {\it Attractive} 4 {\it Bags Under Eyes} 5 {\color{violet}{\it Bald}} 6 {\color{green}{\it Bangs}} 7 {\it Big Lips} 8 {\it Big Nose} 9 {\color{blue}{\it Black Hair}} 10 {\it Blond Hair} 11 {\it Blurry} 12 {\it Brown Hair} 13 {\color{red}{\it Bushy Eyebrows}} 14 {\it Chubby} 15 {\it Double Chin} 16 {\it Eyeglasses} 17 {\it Goatee} 18 {\color{blue}{\it Gray Hair}} 19 {\it Heavy Makeup} 20 {\it High Cheekbones} 21 {\color{brown}{\it Male}} 22 {\color{cyan}{\it Mouth Slightly Open}} 23 {\it Mustache} 24 {\it Narrow Eyes} 25 {\color{cyan}{\it No Beard}} 26 {\it Oval Face} 27 {\it Pale Skin} 28 {\it Pointy Nose} 29 {\color{brown}{\it Receding Hairline}} 30 {\it Rosy Cheeks} 31 {\color{green}{\it Sideburns}} 32 {\it Smiling} 33 {\color{violet}{\it Straight Hair}} 34 {\color{violet}{\it Wavy Hair}} 35 {\it Wearing Earrings} 36 {\it Wearing Hat} 37 {\it Wearing Lipstick} 38 {\it Wearing Necklace} 39 {\it Wearing Necktie} 40 {\color{brown}{\it Young}} 
}
\label{derived architecture}
\end{figure}

{\noindent \bf Per-attribute performance.} 
We additionally study the individual attributes in multi-attribute learning. Table \ref{per-attribute} lists the per-attribute error rates of different methods on CelebA Dataset. We compare our GNAS-Shallow-Wide model to the state-of-the-art methods including LANet \cite{liu2015faceattributes}, Inde. \cite{hand2017attributes}, MCNN \cite{hand2017attributes}, M-AUX \cite{hand2017attributes} and PaW \cite{ding2017deep}. Results of the other methods are cited from the corresponding papers. Table \ref{per-attribute} shows that GNAS not only performs the best under metric of average error rate, but also performs the best on 37 of the 40 attributes. Only on attributes of `Attractive', `Oval Face', `Rosy Cheeks', and `Wearing Necktie', GNAS performs equally or a little worse compared to the other methods. It is interesting that these attributes are relatively global facial features, possibly because the tree-structured neural network architecture may be better at modeling local features while be worse at modeling global features. This makes sense as M-AUX model \cite{hand2017attributes} densely connects all of the attributes at the last layer of its neural network, such that the outputs of the global attributes could obtain more high-level semantic information from other local attributes, with the expense of larger model complexity. 

{\noindent \bf Architecture visualization.}
Fig. \ref{derived architecture} shows the network architecture derived by GNAS-Shallow-Thin on LFWA dataset. The neural architecture is tree-structured, where the centering purple node is the root block, and the numbered blue nodes are the 40 attributes. We could find many groupings of attributes which accord with intuition distinctly, and we highlight them with different colors in the caption of Fig. \ref{derived architecture}. For instance, in Fig. \ref{derived architecture}, `Bald', `Straight Hair', and `Wavy Hair' are clearly related. `5'o Clock Shadow', `Arched Eyebrows', and `Bushy Eyebrows' are related to the facial hairs. We also observe that some related attributes are grouped at the lower layers. For instance, `Bangs' and `Sideburns' are hairs around face, and they are grouped with the facial hairs group at the lower layer. These reasonable attribute groupings qualitatively demonstrate the effectiveness of our GNAS.

\section{DISCUSSIONS}
In this paper, we have presented a highly efficient and effective greedy neural architecture search method (GNAS) for the automatic learning of multi-attribute deep network architecture. We have presented reasonable greedy strategies to divide the optimization of global architecture into the optimizations of individual connections step by step, such that the optimal global architecture is composed by the optimal local architectures. 

GNAS is efficient due to its greedy strategies and effective due to its large search space. In experiments, GNAS discovers network architecture on 1 GPU in no more than 2 days to outperform the state-of-the-art multi-attribute learning models with fewer parameters and faster testing speed. Quantitative and qualitative studies have been further conducted to validate the efficacy of GNAS.

GNAS is a universal neural architecture search framework, such that it is able to be applied to tree-structured network with arbitrary NN blocks and connections. We can arbitrarily specify the type of an individual block (e.g., vector, 2D feature map), and the type of an individual connection (e.g., MLP, 1D convolutions, 2D convolutions, or even more complex NN architectures) as long as the shape of that connection is valid between two blocks. In the future study, it is encouraged to develop GNAS to various application scenarios by accommodating different optimization techniques of AutoML.

\section*{ACKNOWLEDGEMENTS}
The authors thank for Hieu Pham's beneficial dicussions.
This work is supported by NSFC (U1509206, 61672456, 61472353, 61751209, and 61772436), the fundamental research funds for central universities in China (2017FZA5007), National Basic Research Program of China (2015CB352302), the Key Program of Zhejiang Province, China (2015C01027), Sichuan Science and Technology Innovation Seedling Fund (2017RZ0015, 2017020), Artificial Intelligence Research Foundation of Baidu Inc., and the funding from HIKVision and ZJU Converging Media Computing Lab.

\bibliographystyle{ACM-Reference-Format}
\bibliography{acmart}

\end{document}